\documentclass{article}
\usepackage[utf8]{inputenc}
\usepackage{xcolor}
\usepackage{graphicx}

\usepackage{algorithm, algorithmic}

\usepackage{authblk}

\title{Graph-based Extreme Feature Selection for Multi-class Classification Tasks}

\author[1]{Shir Friedman }
\author[2]{Gonen Singer}
\author[1]{Neta Rabin}
\affil[1]{Tel-Aviv University, Dept. of Industrial Engineering, Tel-Aviv, Israel }
\affil[2]{Bar-Ilan University, Faculty of Engineering, Ramat-Gan, 52900, Israel }

\begin{document}
\date{}
\maketitle

\begin{abstract}

When processing high-dimensional datasets, a common pre-processing step is feature selection.  Filter-based feature selection algorithms are not tailored to a specific
classification method, but rather rank the relevance of each feature with respect to the target and the task. This work focuses on a graph-based, filter feature selection  method that is suited for multi-class classifications tasks. We aim to drastically reduce the number of selected features, in order to create a sketch of the original data that codes valuable information for the classification task. The proposed graph-based algorithm is constructed by combing the Jeffries-Matusita  distance with a non-linear dimension reduction method, diffusion maps. Feature elimination is performed based on the distribution of the features in the low-dimensional space. Then, a very small number of feature that have complementary separation strengths, are selected. Moreover, the low-dimensional embedding allows to visualize the feature space. Experimental results are provided for public datasets and compared with known filter-based feature selection techniques.

\end{abstract}
\section{Introduction}
\label{introduction}
Generating and storing large amount of data for improving automatic learning tasks have become common practice in the industry. However, with storage becoming more and more expansive, companies will eventually have to  limit the amount of stored data, and select subsets of data according to a cost-benefit policy \cite{pugh2011getting}.
Feature selection techniques, which where initially focused on  improving the learning accuracy of models in case of redundant features, and on reducing the computational complexity, may now become valuable for reducing storage space of data warehouses.  Hence, as we enter the  era of big data overflow, the importance of generating sketches of large dataset, by selecting the most relevant features, is clear.  

Feature selection algorithms can be coarsely separated into three types: wrapper, embedded and filter methods \cite{venkatesh2019review}.  Wrapper  methods  perform an iterative search for finding the best subset of features  \cite{jovic2015review}. The selection procedure in is combined with a specific machine learning algorithm, and the selected subset of features is aimed at optimizing the performance of this specific machine learning algorithm.  The main drawback is that is in the case of  high-dimensional data sets, wrapper methods may have a high  computational complexity \cite{manosij2020wrapper}. 

In embedded method, the  features are  selected during the learning process, thus requiring to modify the machine learning algorithm \cite{monteiro2011embedded}. 
Although these methods provide efficient solutions for classification tasks, they depend  on the specific classifier, which may be an obstacle for industry users that apply off-the-shelf classification methods, who will prefer separated feature selection algorithm for the classification algorithms. 

Filter-based feature selection algorithms are not tailored to a specific
classification method, but rather rank the relevance of each feature with respect to the target and the task. Relief \cite{kira1992feature} and ReliefF \cite{robnik2003theoretical} are a well-known filter method, which ranks the quality of a features based on their ability to distinguish between data instances that are close to one another. However, these methods does not remove redundant features. 
In the Fisher score method \cite{hart2000pattern}, the feature are ranked and selected independently according to their  Fisher criterion scores, which is the ratio of inter-class separation and intra-class variance. 
Correlation-based feature selection (CFS , \cite{hall1999correlation})
 selects features that have low linear relations with other features, but are correlated with the label. A feature selection method that is based on mutual information select a subset of features that minimize  redundancy  maximize  new  information was proposed in \cite{wang2017feature}. Maximum Relevance Minimum Redundancy (MRMR \cite{peng2005feature}) finds a minimal optimal subset of features. It resembles ideas that are presented in this work, although the algorithmic approach is different and its iterative nature  may be a drawback when the number of features is large.

In recent years, graph-based methods have been proposed for filter feature selection. The main idea is to construct a feature-graph for modeling the feature space, and then evoke graph-based algorithms for selection of a representative subset of features. The graph allows to code the pairwise relationships between the features, taking into account both relevance and redundancy. A hybrid feature selection approach that is based on Feature Association Map  was developed in \cite{das2017new}. Selection of the best feature subset relays on graph-theoretic principles,  minimal vertex cover and maximal independent set. The INF-FS approach \cite{roffo2020infinite} for the supervised case, constructs a feature-graph in which the edges take into account several metrics such as Fisher criterion, normalized mutual information and the feature's normalized standard deviation. A single rank score is calculated by letting the paths on the graph go to infinite. A graph-based feature selection method that deals with the case of multi-label data, was developed in \cite{hashemi2020mgfs}. The PageRank algorithm, which operates on the graph, derived the features' ranking. 

In this work, the focus is on feature selection for multi-class classifications tasks. When the label consists of multiple classes, the contribution of each feature should be defined by how well it separated between all pairs of classes. Nevertheless, we would not want to select two {\it strong} features if their separation abilities are similar. In the multi-class case, avoiding redundancy in the selected feature subset, may be achieved by selection of features that are capable of complementary discrimination capabilities. Therefore, our goal is to find a subset of features that do not necessarily hold the best, highest, pairwise separability scores, but rather that their combination {\it covers} the feature space in terms of pairwise class separation abilities. In order to form such a graph, we combine a graph-based manifold learning method named diffusion maps (DM) \cite{coifman2006diffusion} that constructs a compact representation of a dataset while preserving its geometric structure based in the Jeffries-Matusita (JM) distance, denoted by JM-DM. The JM distance used to quantify the separation abilities of each single feature between each two pairs of classes. We note that other distances like the Wasserstein distance may be considered instead of JM. Given a multi-class problem, the JM based feature separation is coded by symmetric matrix that holds the class separability strength of each feature. Application of diffusion maps to the feature space, where each feature is represented by a JM distance matrix, provides a visual description of the feature distribution with respect to class separability. In order to prioritize features that do not necessarily have the highest averaged separation \cite{khosravi2018msmd}, but contribute to specific class-separation pairs, a simple k-means algorithm is evoked in the low-dimensional space. It results with a sampled subset of features that are associated with different JM matrices, thus they have different separation capabilities. Features with similar JM distance matrices lie close to one another are eliminated. 

Experimental results are demonstrated on two high-dimensional datasets and are compared with known common used filter methods for feature selection.  We show that the proposed JM-DM  procedure results in higher classification accuracy, especially when we consider extreme cases where a very small number of features from the original dataset are kept. 
The paper is organized as follows. Section \ref{sec:methods} describes the JM distance measure and diffusion maps (DM). The proposed feature selection algorithm is described in Section \ref{sec:JM-DM}. Experimental results are presented in Section \ref{sec:results}. Last, conclusions are provided in Section \ref{sec:conclusions}.

\section{Methods}
\label{sec:methods}
This section reviews the Jeffries-Matusita (JM) distance measure and diffusion maps.
Denote the learned dataset by $(X,Y)$.  $X$ is a dataset of size $N \times M$. Here, $N$ is the number of samples and $M$ is the dimension of the feature space. The label, of size $N \times 1$ is stored in the vector $Y$, we assume that there are $C$ classes, meaning that $Y$ contains $C$ different values. The dataset $X$ is comprised of $M$ feature vectors, denoted  by $F = \{f_1,\ldots f_M\}$, each $f_i$ is  of size $1 \times N.$ In general, we assume that $X$ holds at least one instance from each class.

\subsection{Jeffries-Matusita distance for feature selection}\label{sec:JM}

Given a feature $f_i \in F$, the JM distance constructs a $C \times C$ matrix that defines how well the feature $f_i$ separates between {\it all} pairs of classes. This matrix is denoted by $JM_i$. The matrix entry $JM_i(c,\tilde{c})$ informs us how well the feature $f_i$ separates between the two classes $c$ and $\tilde{c}$, where $1 \le c,\tilde{c} \le C$.  The matrix entries are computed by 
\begin{equation}\label{eq:JM}
JM_{i}(c,\tilde{c}) = 2\left(1-e^{-B_{i}(c,\tilde{c})}\right),
\end{equation}
where
\begin{equation}\label{eq:Bh}
B_{i}(c,\tilde{c}) = \frac{1}{8}\left(\mu_{i,c} -\mu_{i,\tilde{c}}\right)^2 \frac{2}{\sigma^2_{i,c} +\sigma^2_{i,\tilde{c}}}
+\frac{1}{2}\mbox{ln}\left(\frac{\sigma^2_{i,c} +\sigma^2_{i,\tilde{c}}}{2\sigma_{i,c} \sigma_{i,\tilde{c}}}\right)
\end{equation}
is the Bhattacharyya distance. The values $\mu_{i,c}, \mu_{i,\tilde{c}}$ and $\sigma_{i,c}, \sigma_{i,\tilde{c}}$ are the mean and variance values of
two given classes $c$ and $\tilde{c}$ from the feature $f_i.$

\subsection{Diffusion maps}\label{sec:DM}
Diffusion maps (DM) allows to  model high-dimensional that lie on a non-linear manifold. Given a dataset ${\bf Z} = \{z_1, z_2, \ldots, z_M\}$ with M data points, 
a graph  ${\bf G} = \left( {\bf Z},{\bf W} \right)$ is constructed, where the points in ${\bf Z}$ are the vertices of ${\bf G}$ and a kernel matrix ${\bf W}$ of size $M \times M$,  holds the graph's weighted edges. ${\bf W}$ should satisfy the following properties: symmetric, positive-preserving,	and positive semi-definite (see \cite{coifman2006diffusion} for details). 

The Gaussian kernel  ${\bf W}= w\left(z_i, z_j\right)=e^{\frac{-\|z_i-z_j\|^2}{2\epsilon}}$
is a common choice for the weight matrix of ${\bf G}$. The scale of the kernel $\epsilon$ defines the local neighborhood around  each data point in the original space. The value of $\epsilon$ should be adapted to the density distribution of the data, and may be set by the following heuristic 
$\epsilon = \max_{j} [\min_{i, i\ne j} (\| z_i - z_j\|)^2 ].$
This value can be further controlled by  a factor  multiplication. We note that the Gaussian kernel may be replaced by other kernels that obay the obove properties.

By introducing a  scale parameter $\alpha$, the effect of the non-uniformed data distribution may be controlled. A general normalized form of the kernel is given by
\begin{equation}
{\bf W}_{\alpha}=w_{\alpha}(z_i,z_j)=\frac{w(z_i,z_j)}{q^{\alpha}(z_i)q^{\alpha}(z_j)}, \;\;\;\; q\left(z_i\right)=\sum_{z_j \in {\bf Z}}w\left(z_i,z_j\right).
\end{equation}
Here, we set $\alpha =1$, which results in an approximation of ${\bf K}$ to the Laplace-Beltrami operator \cite{coifman2006diffusion}, and it allows to recover the geometry of the data points, regardless of their distribution.
A second normalization is performed for generating a Markov transition matrix ${\bf K}$ from ${\bf W}_{\alpha}$. This results in
\begin{equation}
{\bf K} = {\bf D}^{-1}{\bf W}_{\alpha}, \;\;\;\ {\bf D} = d(z_i,z_i) = \sum_{z_j \in {\bf Z }}w_{\alpha}\left(z_i,z_j\right).
\end{equation}

The spectral decomposition  of ${\bf K}$ yeilds a set of embedding coordinates for the dataset ${\bf Z}.$ Denote the eigenvalues of ${\bf K}$ by $\{\lambda_{l}\}_{l=0}^{M-1}$
and the left and right  eigenvectors by $\{\phi_{l}\}_{l=0}^{M-1}$ and $\{\psi_{l}\}_{l=0}^{M-1}$,  respectively. Although  ${\bf K}$ is not a symmetric matrix, it is conjugate to a symmetric matrix, thus the two sets of eigenvectors  $\{\phi_{l}\}_{l=0}^{M-1}$ and  $\{\psi_{l}\}_{l=0}^{M-1}$ , are biorthonormal $\left<\phi_{m},\psi_{l} \right> = \delta_{l,m}$. Therefor, each element in ${\bf K}$ can be computed by
\begin{equation}
\label{eq:K_spectral}
k(z_i,z_j) = \sum_{l=0}^{M-1}\lambda_{l}\psi_{l}(z_i)\phi_{l}(z_j).
\end{equation}

The  matrix ${\bf K}$ has a decaying spectrum, $\lambda_{l} \rightarrow 0$
as $l$ grows. This allows to approximate the entries of ${\bf K}$ in Eq.  (\ref{eq:K_spectral})  by considering  a small number of terms $d$ in the sum.  Finally, the diffusion maps coordinates are defined by
\begin{equation}
\label{eq:DM_cords}
{\bf \Psi}(z_i) = \left(
\lambda_{1}\psi_{1}(z_i) ,
\lambda_{2}\psi_{2}(z_i) ,
\lambda_{3}\psi_{3}(z_i) , \cdots
\right).
\end{equation}
The first $d \le M$ diffusion maps coordinates are considered, thus the space is reduced. Algorithm \ref{alg:DM} summarizes the DM construction.

\begin{algorithm}[H]
\caption{Diffusion maps}
\textbf{Input:} Dataset  ${\bf Z}=\{z_1,\ldots,z_M\}$. Kernel scale parameter $\epsilon$.  \\
\textbf{Output:} Diffusion maps embedding coordinates ${\bf \Psi}(z_i)$.\\
 \begin{algorithmic}[1]
 \label{alg:DM}

    \STATE Construct the Gaussian kernel ${\bf W}_{ij}= w(z_i, z_j) = e^{\frac{-\|z_i-z_j\|^2}{2\epsilon}}.$

    \STATE Normalize ${\bf W}$ by ${\bf W}_{\alpha}=\frac{w(z_i,z_j)}{q(z_i)q(z_j)},$  where $q\left(z_i\right)=\sum_{z_j}w\left(z_i,z_j\right).$

    \STATE  Normalize   ${\bf W}_{\alpha}$ to be a Markov transition matrix ${\bf K} = {\bf D}^{-1}{\bf W}_{\alpha},$ 
    where ${\bf D}$ is a diagonal matrix with ${\bf D} = d(z_i, z_i) = \sum_{z_j}w_{\alpha}\left(z_i,z_j\right).$

\STATE Compute the spectral decomposition of {\bf K} by \\
$k(z_i,z_j) = \sum_{l=0}^{M-1}\lambda_{l}\psi_{l}(z_i)\phi_{l}(z_j).$

    \STATE  Construct  diffusion maps coordinates ${ \bf \Psi}(z_i) = \left(\lambda_{1}\psi_{1}(z_i) , \lambda_{2}\psi_{2}(z_i) , \cdots \right).$

 \end{algorithmic}
\end{algorithm}

\subsubsection{Diffusion Distances}
\label{sec:diff_distances}

Diffusion maps provide a locally preserving metric
named diffusion distance that is defined on the data points. Two data points $z_i$ and $z_j$ are close in this metric if the graph $G$ holds many paths that connect between the two points. In this work, two features with similar JM matrices are guarantied to be embedded close to one another, making DM a suitable method to fulfil our objectives.  
Mathematically, the diffusion distance is defined by 
\begin{equation}
\label{eq:diff_dist}
    D^2(z_i,z_j) = \sum_{z_m \in Z}\frac{\left(k(z_i,z_m) - k(z_m,z_j)\right)^2}{\phi_0(z_m)},
\end{equation}
where $\phi_0$ is the first left eigenvector of ${\bf K}$. The spectral decomposition of ${\bf K}$, as defined in Eq. (\ref{eq:K_spectral}) is substituted into the  nominator of Eq. (\ref{eq:diff_dist}) to yield the diffusion distance expressed by the DM embedding coordinates, which is expressed by

\begin{equation}
\label{eq:diff_dist2}
    D^2(z_i,z_j) = \sum_{l} \lambda_l^2 \left(\psi_{l}(z_i) - \psi_{1}(z_j) \right)^2.
\end{equation}
From Eqs. (\ref{eq:diff_dist}) and (\ref{eq:diff_dist2}), we deduce that the Euclidean distance between the data points (here the features)  low-dimensional representation  is equivalent to the random-walk distance  as defined in the original space. This fact gives rise to the idea of selecting the features in the generated low-dimensional space.

\section{JM-DM Feature Selection}
\label{sec:JM-DM}
The proposed feature selection approach consists of three steps.  First, each feature is replaced  with its computed JM matrix. Next, DM is applied to create a low-dimensional embedding of the feature space. Last, a selection procedure is applied for picking a subset of relevant features from the low-dimensional space.

The DM coordinates provide a reduced and compact space that organizes the features by their underlying class separability properties. Using a Gaussian kernel with Euclidean distances will keep features with similar JM matrices close to each other in this space.  Here, we suggest to perform the elimination process by application of k-means on the embedded feature space, where $k^{\ast}$ is the desired number of selected features, as defined by the user. K-means act as a simple sampling mechanism, choosing features from different areas of the embedded space, therefore, these features have different separability characteristics. In order to improve the k-means selection procedure, and to shift the selection such that the worse features (in terms of the JM separability measure) will not be selected, a cut-off parameter $q$ is defined. This cut-off parameter defines the quantile of low-seperability features to be ignored in the selection process. In order to calculate a naive ranking of the features, we use the average JM matrix values.   Algorithm \ref{alg:JMDM} summarizes  the proposed feature selection approach. 

\begin{algorithm}[H]
\caption{JM-DM feature selection}
\textbf{Input:} Feature vectors $F=\{f_1,\ldots,f_M\}$. Number of selected features $k^{\ast}$, Quantile of features to ignore $q$, DM kernel scale parameter $\epsilon$.    \\
\textbf{Output:} A subset of selected features $F_S$. \\
 \begin{algorithmic}[1]
 \label{alg_JM_DM}

    \STATE For each feature $f_i,$ compute the JM-matrix $JM_i$ of size $C \times C$. 
    
    \STATE Compute the average value of each JM-matrix, ${\bf m}(i) = \mbox{average}(JM_i).$  
    
    \STATE Reshape $JM_i$ to be a vector of size $1 \times (C \times C)$ and form a dataset ${\bf Z}$ that holds the reshaped JM matrices as its rows.  
    
    \STATE Apply Algorithm \ref{alg:DM} to ${\bf Z}.$ This results with DM coordinates  \\${\bf \Psi}(JM_i) = \left(\lambda_{1}\psi_{1}(JM_i) , \lambda_{2}\psi_{2}(JM_i) , \lambda_{3}\psi_{3}(JM_i) , \cdots \right).$

    \STATE Sort ${\bf \Psi}(JM_i)$ according to the values of the vector ${\bf m}$ and keep the reduced set ${\bf \Psi}_q(JM_i)$ such that ${\bf m}(i)$ is above the  q-qauntile.
    \STATE Apply k-means to ${\bf \Psi}_q(JM_i)$. From each  of the $k^{\ast}$ clusters select the representative feature as the feature with the highest value of ${\bf m}$.
    
    \STATE Return the list $F_S$ that holds the $k^{\ast}$ selected features. 
    
 \end{algorithmic}
\label{alg:JMDM}
\end{algorithm}

\section{Experimental Results} \label{sec:results}

Two public datasets from the UCI repository \footnote{UCI: https://archive.ics.uci.edu/ml/datasets.php} belonging to  multi-class classification, with a large nuber of features, tasks are considered. 
\begin{enumerate}
    \item{\bf Isolet:} The dataset holds 617 features extracted from voice recordings of 150 subjects, who spoke out each of the English alphabets. The number of instances is $7797$. The task is to classify the correct letter ($26$ classes). 
    \item{ \bf Multiple features (digit recognition):} The dataset consists of $649$ features of handwritten digits $0,1,\ldots,9$. These features include Fourier coefficients, profile correlations,  Karhunen-Love coefficients, pixel averages and morphological features. The number of instances is $2000$, $500$ per class.
\end{enumerate}

To benchmark the proposed method, several filter selection methods, including ReleifF, Fisher, Correlation-based selection (CFS),  Maximum Relevance Minimum Redundancy (MRMR) were applied, while a 5-fold cross validation scheme was used for each data set. The classification measures were averaged over the $5$ folds. In addition, we compared our results to a ``random filter" (Rand. in the results tables), in which the number of desired features are randomly selected from the input feature columns. 
 Classification was carried out using  k-nearest neighbors (KNN, k=5)  and classification trees (denoted by C-TREE in the tables below). The number of DM coordinates in Alg. \ref{alg:DM} was set to $3$ and the value of $q$ in Alg. \ref{alg:JMDM} was set to $0.25$. For each of the datasets we selected $\frac{1}{8}$, $\frac{1}{16}$ and $\frac{1}{32}$ of the original features. These correspond to 77, 39 and 19 features for the Isolet dataset, and 81, 41, and 20 for the multiple feature dataset.

\begin{figure}[h]
\centering
\includegraphics[scale=0.4]{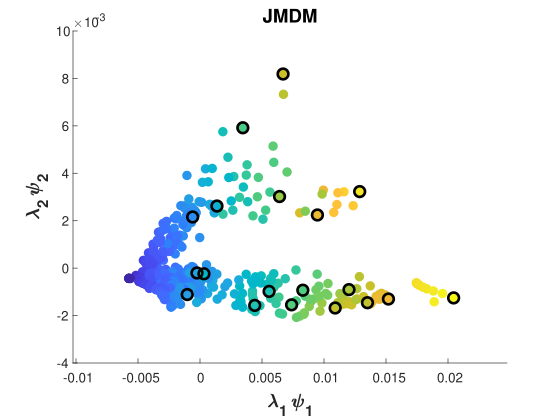}
\includegraphics[scale=0.4]{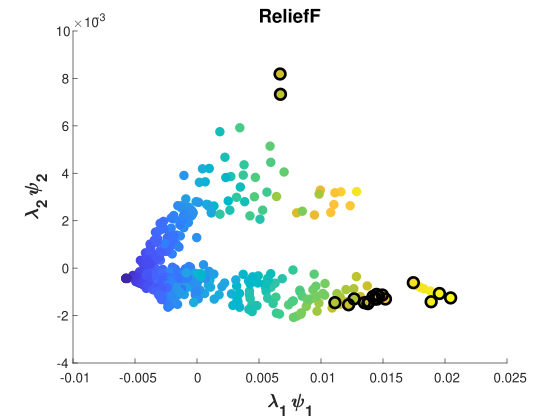}
\caption{DM embedding of the Isolet features, colored by the average JM score. The $\frac{19}{617}$ features that were selected by  JM-DM (left) and by ReleifF (right)  are circled in black. JM-DM  selects features with different separation strengths, while ReleifF selects features with strong average separation abilities. } \label{fig:isolet_DM}
\end{figure}

 Figure \ref{fig:isolet_DM} plots the DM embedding of the ISOLET feature space as well as features  selected by the JM-DM and the ReleifF algorithms. Each feature is colored by the mean value of its computed JM matrix. Features with stronger average separation are colored in  yellow, while features with a weak class separation are colored in blue.

 Table \ref{table:isolet} presents the classification results for the Isolet dataset. For the first case, where the features are reduced by  $\frac{1}{8}$, the results of the proposed JM-DM algorithm are significantly better for all cases, using a paired t-test with p-value of $0.05$. When the proportion of selected features is  $\frac{1}{16}$, the results of a JM-DM are significantly better than  ReleifF, Fisher, CFS and random for the C-TREE classification. Last, when only $\frac{1}{32}$ of the features are kept, JM-DM is significantly better than all of the algorithms except MRMR.

\begin{table}[h]
\caption{Classification Results for the Isolet Dataset (617 input features)}
{\begin{tabular}{|c | c | c | c | c | c | c| c|}
\hline
 \multicolumn{1}{|p{1.6cm}|}{\centering {\bf  Alg.}} & \multicolumn{1}{|p{1.6cm}|}{\centering {\bf Fraction of \\ Selected Features}} & \multicolumn{6}{|p{6cm}|}{\centering {\bf Feature Selection Method}}  \\
{}   & {} &
 {\bf JM-DM} & {\bf ReliefF} & {\bf Fisher} & {\bf CFS} & {\bf MRMR} & {\bf Rand.}\\
\hline
\hline
 KNN &  $1 \slash 8$ & {\bf 0.864} &	0.725 &	0.790 &	0.642 & 0.850	& 0.330 \\ 
 \hline
 C-TREE &  $1 \slash 8$ & {\bf 0.760} &	0.6663 &	0.708 &	0.525 & 0.743 &	0.698 \\  
 \hline
 \hline
  KNN &  $1 \slash 16$  & 0.800	& 0.638 &	0.653 &	0.582 &	{\bf 0.819} & 0.583 \\ 
 \hline
 C-TREE &  $1 \slash 16$ & {\bf 0.719} &	0.586 &	0.597 &	0.483 &	0.717 & 0.648 \\
 \hline
 \hline
  KNN &  $1 \slash 32$ & {\bf 0.733}	& 0.552 &	0.537 &	0.472 &	0.720 & 0.350 \\ 
 \hline
 C-TREE &  $1 \slash 32$ & {\bf 0.666} &	0.494 &	0.477 &	0.422 & 0.649 &	0.506\\ 
 \hline
 \end{tabular}}
\label{table:isolet}
\end{table}

 Table \ref{table:mulfeatures} presents the classification results for the Multiple features dataset. For the first case, where the features are reduced by  $\frac{1}{8}$, the results of the proposed JM-DM algorithm are significant for all cases, expect for CFS and MRMR with KNN classification. When  $\frac{1}{16}$  and  $\frac{1}{32}$ of the features are kept, JM-DM is significantly better than all of the other compared algorithms.  It is surprising to see  random selection of features outperforms some of the tested methods.

\begin{table}[h]
\caption{Classification Results for the Multiple-Features Dataset (649 input features)}
{\begin{tabular}{|c | c | c | c | c | c | c| c|}
\hline
 \multicolumn{1}{|p{1.6cm}|}{\centering {\bf  Alg.}} & \multicolumn{1}{|p{1.6cm}|}{\centering {\bf Fraction of \\ Selected Features}} & \multicolumn{6}{|p{6cm}|}{\centering {\bf Feature Selection Method}}  \\
{}   & {} &
 {\bf JM-DM} & {\bf ReliefF} & {\bf Fisher} & {\bf CFS} & {\bf MRMR} & {\bf Rand.}\\
\hline
\hline
 KNN &  $1 \slash 8$ & 0.975 &	0.961 &	0.840 &	{\bf 0.978} & 0.974	& 0.967 \\ 
 \hline
 C-TREE &  $1 \slash 8$ & {\bf 0.939} &	0.917 &	0.826 &	0.906 & 0.898 &	0.884 \\  
 \hline
 \hline
  KNN &  $1 \slash 16$  & {\bf 0.976}	& 0.892 &	0.786 &	0.962 &	0.951 & 0.954 \\ 
 \hline
 C-TREE &  $1 \slash 16$ & {\bf 0.927} &	0.873 &	0.723 &	0.897 &	0.856 & 0.865 \\
 \hline
 \hline
  KNN &  $1 \slash 32$ & {\bf 0.964}	& 0.831 &	0.825 &	0.905 &	0.892 & 0.896 \\ 
 \hline
 C-TREE &  $1 \slash 32$ & {\bf 0.919} &	0.830 &	0.733 &	0.845 & 0.780 &	0.834\\ 
 \hline
 \end{tabular}}
\label{table:mulfeatures}
\end{table}

\section{Conclusions}
In this paper, we proposed a feature selection algorithm that is suited for multi-class classification tasks. Its main idea is to select an extremely small subset of features that have 
complementary separation abilities. This subset does not necessarily hold the features that have the best average separation ability, but rather aims to select features that compensate for one another in terms of differentiating between all pairs of classes.
The algorithm is composed of three  techniques, JM distance matrices, diffusion maps and k-means, which are applied in a sequential manner. The feature space was represented by JM matrices that efficiently and simply code the separation ability of a feature in the case of multi-class problems. Then, in order to retain this detailed information in the selection process, a feature graph was constructed based on the JM matrices. Application of diffusion maps (DM) resulted in a small number of embedding coordinates that organize the feature space based on the JM matrices. Features with similar JM matrices are embedded close to one another in the DM embedding. Last, selection of embedded features that correspond to different JM matrices, was simply done by application of the k-mean algorithm, where k was the desired number of feature to preserve.  The experimental results highlighted the advantage of the proposed approach, as it achieved higher classification results compared to classical feature selection techniques. 
The proposed algorithm may be viewed as a more general framework, such that other metrics, besides JM, may be used and explored in the future. Another direction can explore the effect of fusion of several separation metrics as a basis for construction of the features-graph, which is the input to the DM algorithm.
\label{sec:conclusions}


\end{document}